\documentclass[conference]{IEEEtran}
\IEEEoverridecommandlockouts
\usepackage{cite}
\usepackage{amsmath,amssymb,amsfonts}
\usepackage{algorithmic}
\usepackage{graphicx}
\usepackage{textcomp}
\usepackage{xcolor}
\usepackage{booktabs}
\def\BibTeX{{\rm B\kern-.05em{\sc i\kern-.025em b}\kern-.08em
    T\kern-.1667em\lower.7ex\hbox{E}\kern-.125emX}}
\usepackage{multirow}
    
\begin{document}

\title{Online Topological Localization\\ for Navigation Assistance in Bronchoscopy }
\author{\IEEEauthorblockN{Clara Tomasini, Luis Riazuelo, Ana C. Murillo}
\IEEEauthorblockA{\textit{DIIS - I3A - Universidad de Zaragoza} \\
Zaragoza, Spain \\
ctomasini@unizar.es}
}

\maketitle

 \begin{abstract}
Video bronchoscopy is a fundamental procedure in respiratory medicine, where medical experts navigate through the bronchial tree of a patient to diagnose or operate the patient. Surgeons need to determine the position of the scope as they go through the airway until they reach the area of interest. 
This task is very challenging for practitioners due to the complex bronchial tree structure and varying doctor experience and training. Navigation assistance to locate the bronchoscope during the procedure can improve its outcome.  
Currently used techniques for navigational guidance commonly rely on previous CT scans of the patient to obtain a 3D model of the airway, followed by tracking of the scope with additional sensors or image registration. These methods obtain accurate locations but imply additional setup, scans and training.
Accurate metric localization is not always required, and a topological localization with regard to a generic airway model can often suffice to assist the surgeon with navigation. We present an image-based bronchoscopy topological localization pipeline to provide navigation assistance during the procedure, with no need of patient CT scan.
Our approach is trained only on phantom data, eliminating the high cost of real data labeling, and presents good generalization capabilities. The results obtained surpass existing methods, particularly on real data test sequences.
\end{abstract}

\begin{IEEEkeywords}
localization, segmentation, bronchoscopy
\end{IEEEkeywords}
\section{Introduction}
 Video Bronchoscopy is a fundamental diagnostic and therapeutic procedure for respiratory medicine. During a bronchoscopy procedure, doctors explore the airway with a camera to detect potential issues. It is frequently used for biopsies, object retrieval, or diagnosis and assessment of evolution of conditions such as cancer lesions. For example, it is crucial in the case of lung cancer detection, as the diagnosis can only be obtained through biopsy of the affected tissue. 

\begin{figure}
    \centering
    \includegraphics[width=0.95\linewidth]{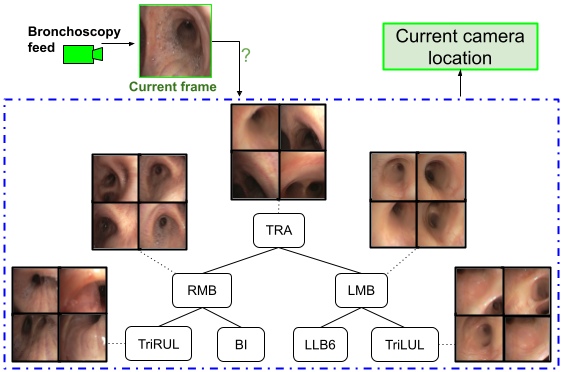}
    \caption{Our pipeline determines the topological location of the camera, given the video feed observed up to the current frame (in green), with regard to a generic tree model of the airway (with nodes named as TRA, RMB, LMB,..., as done clinically).  Examples of frames for some of the classes are shown in the diagram, highlighting the low inter-class variation. 
    }
    \label{fig:intro}
\end{figure}

Traditionally, once the location of a region of interest such as a lesion has been identified, for example through previous CT scans, doctors try to reach that region by mentally keeping track of the bronchoscope localization while manually navigating through the airway. However, this poses a real challenge due to the airway's complex, tree-like structure and lack of visual differences between the different segments (as shown in Fig.~\ref{fig:intro}), as well as varying training and experience levels of the doctors. Potential airway segments incorrect identification might lead to longer procedures for the patient and possibly missed lesions. Navigational assistance can be beneficial in reducing the error rate~\cite{merritt2008image,memoli2012meta} and intervention time for the patient~\cite{shinagawa2007virtual}. 

Currently, navigation guidance can be obtained through Electromagnetic Navigation Bronchoscopy (ENB)~\cite{reynisson2014navigated} and Virtual Navigation Bronchoscopy (VNB)~\cite{ferguson2005virtual}. Both methods rely on rendering previous CT scans of the patient in order to obtain a 3D reconstruction of the airway. The path to follow to reach a specific point is planned prior to the procedure on the reconstructed airway. Then, during the procedure, ENB tracks the bronchoscope through that model using an additional electromagnetic scope placed on the bronchoscope during the procedure~\cite{franz2014electromagnetic}. In VNB, tracking of the camera can be done through image registration~\cite{mori2004new,rai2008combined} of the previously rendered images and the real images captured by the camera.  However, both methods require additional equipment, training, planning time and pre-operative CT scans of the patient. 

To make VNB more accurate and accessible, more recent approaches propose to use SLAM-based pipelines~\cite{wang2020visual,visentini2017deep}. Other works perform registration of the RGB images captured by the scope with the rendered CT images using depth estimation~\cite{shen2019context,banach2021visually}. The generated depth map for each RGB frame is registered to the CT depth map to obtain pose estimation. However, obtaining correct registration between CT-rendered and RGB images is still challenging due to image characteristics, possible occlusions or specific bronchoscopy lighting conditions.  

\begin{figure*}[!htb]
    \centering
    \includegraphics[width=0.82\linewidth]{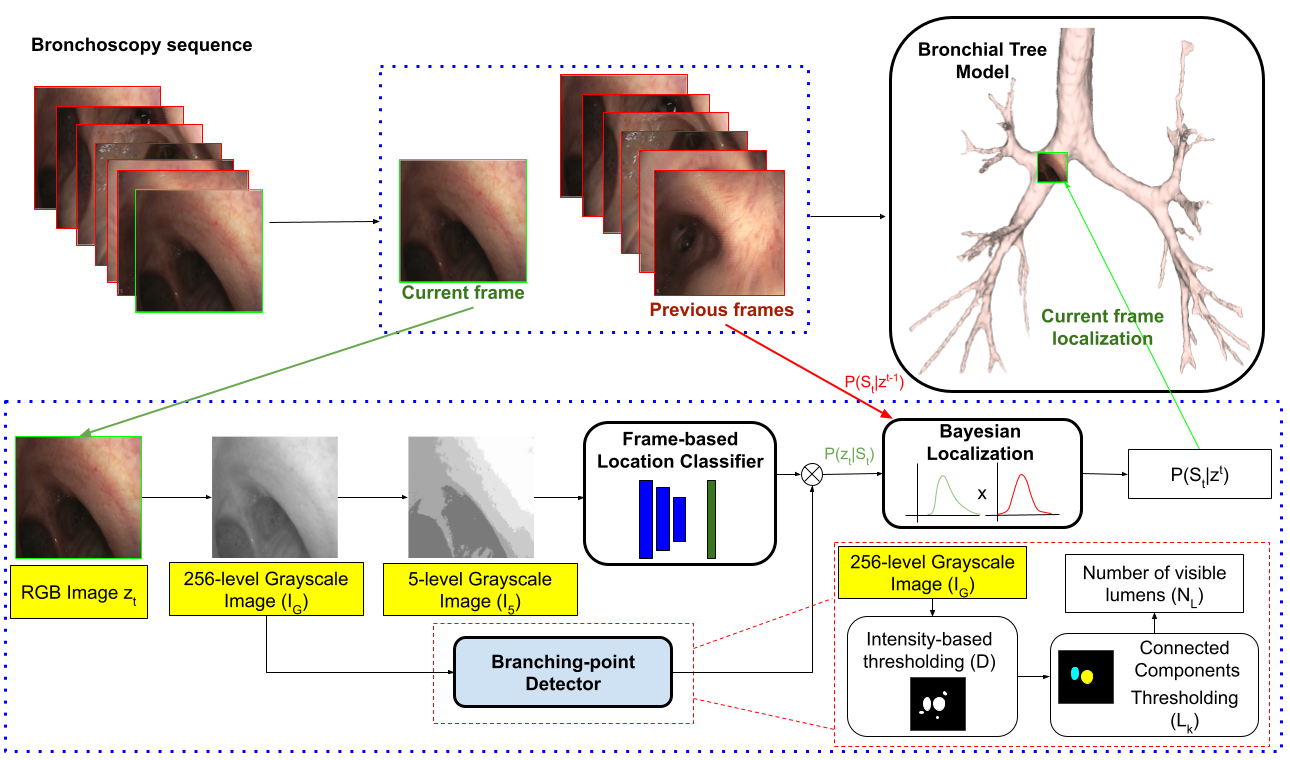}
    \caption{Overview of our proposed bronchoscopy topological localization approach. The current frame in the sequence goes through the frame-based CNN classifier and the branching-point detector. The output is then combined with the distributions of the previous frames to obtain the posterior distribution and the current frame localization in the bronchial tree.}
    \label{fig:approach}
\end{figure*}

While these methods yield good results, the benefits of obtaining an accurate metric localization might not always outweigh the additional setup and time required. A topological localization could be sufficient to provide navigational assistance during bronchoscopy procedures, as well as be useful in possible future autonomous procedures. As such, we present a \textbf{novel online image-based topological localization approach for bronchoscopy} (Fig.~\ref{fig:approach}). Our method relies on a generic tree model of the airway instead of a patient specific, pre-operative CT scan, removing the need for image registration. It is composed of a frame location classifier, trained only on phantom data but generalizing well to real data, and a Bayesian Localization filter. A key ingredient is the \textbf{branching-point detector}, that makes the Bayesian filter more robust to the noise in the classifier. We train our pipeline on phantom data, and evaluate on both phantom data and real data. We show significant performance improvements in real data compared to other methods without relying on any labeled real data.

\section{Related Work}
 
\subsection{Image-based online localization}
A recently proposed approach~\cite{gil2020intraoperative} to image-based online localization is to dynamically create and update an exploration tree to represent the nodes explored along the procedures.
In each frame, ellipses are fitted to the lumen regions or openings visible in the frames~\cite{esteban2016stable}. The ellipses detected are tracked from one frame to the next and a local hierarchy of ellipses is created for the current frame. The nodes in the local hierarchy are matched to the current global on-line tree and the on-line global tree is updated accordingly. The exploration path followed during the procedure is obtained from that global tree after its last update at the last frame. The main downside to this approach is that it creates an abstract airway representation and abstract path. That path only represents nodes, not specific bronchial points, meaning it gives the correct branch level, but doesn't differentiate between the specific nodes at each level. For example, following the bronchial tree model in Fig.~\ref{fig:intro}, the proposed approach doesn't differentiate between RMB and LMB. The direct applicability and benefit of this method to real scenarios is therefore limited.
A way to solve this issue is to match anatomical landmarks in the image such as lumen centers in both frame and patient-specific CT~\cite{sanchez2016navigation}, as is already done for Virtual Bronchoscopic Navigation. However, this requires a corresponding patient CT scan, which implies additional ressources, time and medical imaging for the patient.\\

\subsection{Frame-based location classification} 
Recent works have been presented that propose to localize the bronchoscope through classifying each frame in the sequence individually within the bronchial tree.
In~\cite{yoo2021deep}, various supervised CNN models, such as ResNet50v2~\cite{he2016identity} and DenseNet~\cite{huang2017densely}, are used and show promising results for classifying real bronchoscopy images in the initial levels of the tree. However, this approach classifies each frame individually and does not take into account the temporal aspect of a bronchoscopy procedure and its influence on the localization of scope. Additionally, neither the trained models nor the large set of labeled real bronchoscopy images are publicly available up to our knowledge. Similarly, frame-based classification is used as the first step of a topological localization pipeline in ~\cite{keuth2024airway}, with further temporal filtering to refine this classification based on the rest of the bronchoscopy sequence. 

\subsection{Topological Localization in endoscopy images}
Recently, \cite{keuth2024airway} was proposed as an offline topological localization pipeline in bronchoscopy. 
This method predicts the bronchoscope location within a generic, inter-patient model of the airway, as the airway structure has been proven similar enough up to the fourth branch level to do so~\cite{smith2018human}. It is trained and evaluated on phantom data, for which localization as well as depth map ground-truth are available. The first step of the pipeline is a segmentation network, trained with pseudo-labels obtained using k-means applied to the depth-maps~\cite{keuth2023weakly}, that extracts features from the image. These features are then fed to the classifier to obtain a likelihood for each class. The second step is a Hidden Markov Model (HMM) that adds temporal information and determines the path followed during the procedure. For each frame, the likelihood from the classifier is combined with anatomical constraints from the bronchial tree model, such as the procedure having to begin and end in the trachea (node TRA). 
Though it shows good results in phantom data, it has not been tested in real data. It is also an offline localization pipeline, and could not be used online, for navigational guidance during the procedure, due to the constraints and formulation of the HMM. \\
Another approach, ColonMapper~\cite{morlana2024colonmapper}, was proposed as an image-based pipeline for online topological localization in another type of endoscopy images: colonoscopy. Differently from bronchoscopy, as the colon does not have a specific structure, ColonMapper first builds a topological map of the colon by matching and grouping similar images from one sequence into nodes using an extracted global descriptor for each frame. Next, a Bayesian Localization filter is applied to compute the probability distribution for each image in another sequence. The full posterior distribution for each image is computed using similarity between the global descriptor of that image and each nodes' descriptors as likelihood. A time evolution model adds colonoscopy priors by considering colonoscopies as linear, meaning consecutive frames should jump to close nodes within certain distance.  

\section{Methodology.}
 
\subsection{Overview.}
We present a visual topological localization approach for the bronchial tree with the following key features: 1) the localization of the scope is done with regard to a generic bronchial tree model (avoiding the need of a recent CT scan of the patient), 2) the approach is trained only on phantom RGB data but it is able to directly perform well on real data, 3) the approach can run online.

Following related works \cite{keuth2024airway, morlana2024colonmapper}, we use a pipeline that consists of a \textbf{Frame-based location classifier} followed by a \textbf{Bayesian Localization} filter. This pipeline is illustrated in Fig.~\ref{fig:approach}. 

\begin{figure*}[!htb]
    \centering
    \setlength\tabcolsep{1.5pt}
    \begin{tabular}{r cccc c cccc}
        & \multicolumn{4}{c}{\textbf{Phantom}} &  & \multicolumn{4}{c}{\textbf{Real}} \\
        (a) & \includegraphics[width=0.1\linewidth]{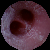} & \includegraphics[width=0.1\linewidth]{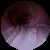} & 
        \includegraphics[width=0.1\linewidth]{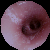} & \includegraphics[width=0.1\linewidth]{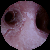} & &
        \includegraphics[width=0.1\linewidth]{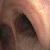} & 
        \includegraphics[width=0.1\linewidth]{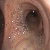} &
        \includegraphics[width=0.1\linewidth]{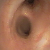} & 
        \includegraphics[width=0.1\linewidth]{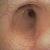}
        \\
        (b) & \includegraphics[width=0.1\linewidth]{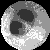} & \includegraphics[width=0.1\linewidth]{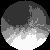} &
         \includegraphics[width=0.1\linewidth]{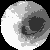} & \includegraphics[width=0.1\linewidth]{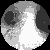} & &
         \includegraphics[width=0.1\linewidth]{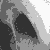} & \includegraphics[width=0.1\linewidth]{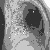} &
         \includegraphics[width=0.1\linewidth]{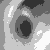} &
         \includegraphics[width=0.1\linewidth]{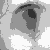}

    \end{tabular}
    
    \caption{Examples of data used: images from a phantom~\cite{visentini2017deep} and real data. (a) Original RGB data as captured during the procedure. (b) 5-level grayscale image, obtained running k-means (k=5) on the RGB image, used as input for our frame-based classifier.}
    \label{fig:train_imgs}
\end{figure*}

Since the texture of the training phantom images is noticeably different from real images (see Fig.~\ref{fig:train_imgs}), our  classifier is trained on  grayscale images (5 levels) instead of RGB images to ignore fine grained details of that texture. Our proposed Bayesian localization approach takes into account the prior from the bronchial tree model and determines when to incorporate the likelihood from the frame-based location classifier with the \textbf{Branching-point detector} proposed. 

\subsection{Bronchial tree model.}
Following~\cite{keuth2024airway}, we use a generic bronchial tree model as a reference to localize the scope. This avoids the need of a the pre-operative CT scan of the patient's airway that most existing methods present. 

The \textit{phantom data} used in \cite{keuth2024airway} considers  a simplified bronchial tree that accounts only for 15 bronchial nodes, up to the fourth generation of the complete bronchial tree. These nodes correspond to those seen in the publicly available phantom data~\cite{visentini2017deep}. This is the same data used for training in our approach. 

The \textit{real data} we introduce for evaluation only explores up to the initial branching tree levels, in particular those depicted in 
Fig.\ref{fig:intro}. These are the tree levels reached in the evaluation sequences.

\subsection{Frame-based location classifier.}
We perform the frame-based location classification using the same network architecture as~\cite{keuth2024airway}, a ResNet model with 800k parameters and 15 classes. Differently from~\cite{keuth2024airway}, we use a 5-level grayscale image as input to the network. To generate this 5-level grayscale image, we convert the original RGB bronchoscopy image to grayscale and apply the K-Means algorithm to discretize it to only 5 gray levels. The resulting image, $I_5$, maintains information on the main structures visible in the frame, but contains less  texture specific information that hinders generalization from phantom to real images. This change allows for better performance in real images even when the model is only trained on phantom images. Examples of frames used as input to the classification model as well as their corresponding original RGB images, from both real and phantom data, can be seen in Fig.~\ref{fig:train_imgs}.

\subsection{Branching-point Detector.}
The frame based location classifier has a high confusion rate, as discussed in the experiments, due to the repetitive nature of the scenarios. To achieve more robust localization results, we only incorporate the information from the frame-based classifier at frames corresponding to branching points, with more discriminative configuration and where the localization is most likely to be close to a change from one node to another. To detect when the scope is at a branching point, the visible lumens are detected as follows: 
\begin{enumerate}
\item \textit{Intensity thresholding}. The grayscale image is filtered to only keep the set of darkest pixels, $D$
\begin{eqnarray}
D = \{ (i,j) | I_G(i,j) < th_I\},
\label{eq:thresholding}
\end{eqnarray}    
\noindent being $I_G$ the gray scale image and $i,j$ the pixel coordinates. In our experiments, we set the intensity threshold $th_I$ to the value corresponding to the 10th percentile (i.e., darkest 10\% of the image intensities).
\item \textit{Connected components}. $D$-pixels are grouped into separate instances $L$ applying a connected components algorithm. The instances are then filtered by area to keep only the largest ones and obtain the number of visible lumens, $N_L$ 
\begin{eqnarray}
    N_L = \sum_k Area(L_k)\geq th_A,%
\end{eqnarray}
\label{eq:thresholding_area}
being $L_k$ each of the separate instances computed. In our experiments, we set the area threshold $th_A$ to 1\% of the image resolution. The scope is considered at a branching-point if $N_L \geq 2$, meaning 2 or more lumens are detected.
\end{enumerate}

\begin{figure}[!htb]
    \centering
    \setlength\tabcolsep{1.5pt}
    \begin{tabular}{ccc}
        \includegraphics[width=0.18\linewidth]{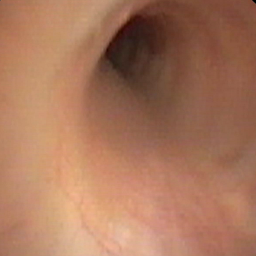} & \includegraphics[width=0.18\linewidth]{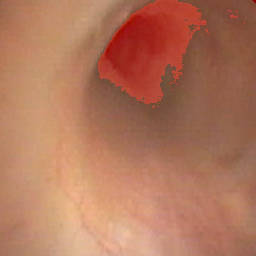} &
        \includegraphics[width=0.18\linewidth]{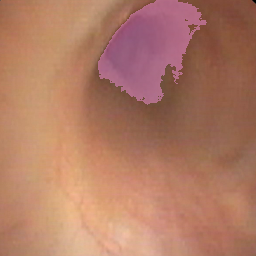} \\
        1-lumen & \multicolumn{2}{c}{1 lumen detected} \\

        \includegraphics[width=0.18\linewidth]{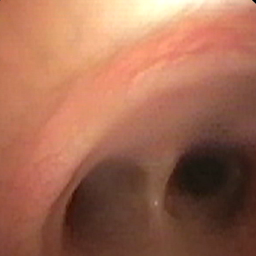} & \includegraphics[width=0.18\linewidth]{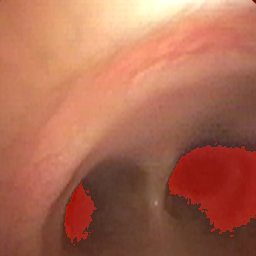} &
        \includegraphics[width=0.18\linewidth]{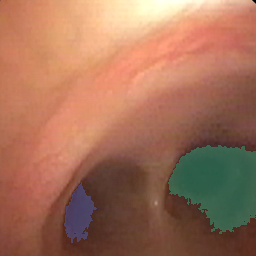} \\
        2-lumens & \multicolumn{2}{c}{2 lumens detected} \\
        
        \includegraphics[width=0.18\linewidth]{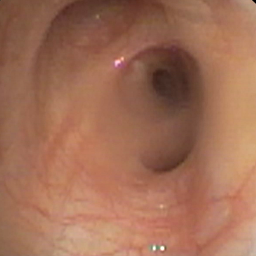} & \includegraphics[width=0.18\linewidth]{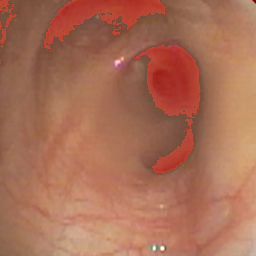} &
        \includegraphics[width=0.18\linewidth]{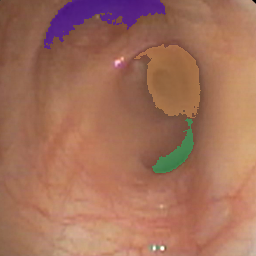} \\
        3-lumens & \multicolumn{2}{c}{3 lumens detected} \\
        
        (a) & (b) & (c)\\      
    \end{tabular}

    \caption{Branching-point detector applied on three real examples: (a) input real image (b) result of intensity threhsolding to select the darkest pixel set $D$ (in red).
    (c) result of applying the connected components algorithm to group $D$-pixels into instances (each color segment represents a different instance $L_k$). The instances are filtered to keep only the largest ones, that correspond to lumen regions indicating different airway channels. }
    \label{fig:detector}
\end{figure}

\noindent
Fig.~\ref{fig:detector} shows the process of our applying these steps to real bronchoscopy images.

\subsection{Bayesian Localization.}
We use a Bayesian Localization filter~\cite{thrun2005probabilistic} to take into account the temporal aspect of the sequence frames seen during the bronchoscopy procedure.
Given frame $z_t$ seen at instant $t$, with $z^t=\{z_0,...,z_t \}$ the frames seen until instant $t$ and $n$ the number of nodes in the bronchial tree, the posterior distribution $p(S_t=i|z^{t}), i \in \{0,...,n\}$, for the localization $S_t$ of the scope within the bronchial tree at instant $t$, is obtained as follows. 
First, the prediction step, Eq.~(\ref{eq:prediction}) , combines the bronchoscopy specific prior $p(S_t=i|S_{t-1}=j)$, defined in Eq.~(\ref{eq:prior1}), with the previous estimations. The bronchoscopy prior takes into account the distance $d$ between each node, obtained from the tree model, and penalizes transitions of over $m$ nodes between two consecutive frames using parameter $\alpha$. In practice, we use $m=1$ and $\alpha=1e-9$, and following related works we consider the distance between consecutive nodes in the tree model to be 1. The prior is normalized before use. 

The location update step, Eq.~(\ref{eq:update}), is only performed if the branching-point detector indicates 2 or more lumens visible in the image. In this case, the likelihood $p(z_t|S_t)$ from the frame-based classifier is taken into account. Otherwise, the likelihood is omitted and we keep the result of the prediction step. Parameter $\eta$ is used to normalize the final posterior.
\begin{equation}
\label{eq:update}
    p(S_t|z^{t}) = \eta p(z_t|S_t)p(S_t|z^{t-1})%
\end{equation}
\begin{equation}
\label{eq:prediction}
    p(S_t|z^{t-1}) = \sum_{j=1}^{n} p(S_t|S_{t-1}=j)p(S_{t-1}=j|z^{t-1})%
\end{equation}
\begin{equation}\label{eq:prior1}
    p(S_t=i|S_{t-1}=j) = \begin{cases}
(1-\alpha*(d(i,j)+1)) & \text{\small if d(i,j)}\leqslant \text{\small m}\\
\alpha*(d(i,j)+1) & \text{\small otherwise}
\end{cases}
\end{equation}
As additional bronchoscopy prior, we consider that a procedure always begins in the trachea (node TRA in the tree), and initialize the posterior distribution for the first frame $P(S_0|z_0)$ with the one-hot vector encoding for the class TRA.

\section{Experiments.}
 
\subsection{Data.}
We train our proposed frame-based location classifier on phantom data containing complete bronchoscopy sequences. Evaluation of our pipeline is done on both real and phantom data sequences. \\
\textbf{Phantom (Ph) data.}
We use the publicly available phantom data from \cite{visentini2017deep}. These sequences are obtained using a phantom model of the airway with the same structure as that of our bronchial tree model (see Fig.~\ref{fig:intro}). The distribution of sequences for training and testing used in our work, as well as corresponding ground-truth localization labels for each frame, are the same as \cite{keuth2024airway}.
\\
\textbf{Real data.}
We evaluate our approach on 2 sequences acquired during real medical practice, with resolution $256\times256$. These sequences have been labeled by experts, and go from the trachea (TRA) up to nodes RMB (sequence 0, 329 frames) and TriLUL (sequence 1, 525 frames) respectively.

\subsection{Experimental setup and Evaluation Metrics.}

We evaluate the topological localization performance using Top-1 and Top-3 Accuracy metrics. 

\textit{Our frame-based classifier} is trained for 30 epochs with learning rate $1e-3$, Adam optimizer and batch-size 32. Data augmentation is done following guidelines in~\cite{cubuk2019autoaugment}.

Our \textit{baseline}, for reference, 
is the frame-based location classifier originally proposed in~\cite{keuth2024airway}.
To allow a fair comparison, we have implemented the complete offline pipeline following the paper (since no code is currently available to run). We train and evaluate the pipeline on the same labeled phantom data used by the authors. Table~\ref{tab:results_previous} presents the Top-1 and Top-3 accuracy reported by the authors and using our re-implementation, for both their proposed frame-based location classifier (\textit{Frame-clf}) and their temporal aware offline localization strategy based on a Hidden Markov Model (\textit{Frame-clf+HMM}). We obtain similar or slightly improved results to those originally reported, confirming that our re-implementation is fair. 

\begin{table}[ht]
\caption{\textbf{offline topological localization} results following~\cite{keuth2024airway} on  sequence 0 of Phantom dataset (\textit{Ph.0}), and sequence 7 of Phantom dataset (\textit{Ph.7}). Top-1 and Top-3 accuracy (\textit{Acc.}) as reported by the authors~\cite{keuth2024airway}, and as obtained with our re-implementation (\textit{Actual}).}
\label{tab:results_previous}
\centering
\begin{tabular}{c c cc cc }
\toprule
& & \multicolumn{2}{c}{\textbf{Top-1 Acc.}} & \multicolumn{2}{c}{\textbf{Top-3 Acc.}} \\

\textbf{Seq.} & \textbf{Method} &  \textbf{\cite{keuth2024airway}} & \textbf{Actual} & \textbf{\cite{keuth2024airway}} & \textbf{Actual} \\ 
\midrule
\multirow{2}{*}{Ph.0} & Frame-clf & 0.81 & 0.78 & 0.88 & 0.93 \\ 
 & Frame-clf+HMM & 0.98 & 0.97 & 1.00 & 1.00 \\
\midrule
\multirow{2}{*}{Ph.7} & Frame-clf & 0.62 & 0.63 & 0.76 & 0.84 \\ 
 & Frame-clf+HMM & 0.66 & 0.84 & 1.00 & 1.00 \\ 
\midrule
\multirow{2}{*}{Ph.(Mean)} & Frame-clf & 0.71 & 0.71 & 0.82 & 0.89 \\ 
 & Frame-clf+HMM & 0.82 & 0.88 & 1.00 & 1.00 \\ 
\bottomrule
\end{tabular}
\end{table}

\subsection{Evaluation of our proposed online pipeline.}
We train our proposed pipeline on phantom data and evaluate it on both phantom and real data. We compare different variations of our pipeline, when using either the baseline frame-based location classifier~\cite{keuth2024airway} or our new frame-based classifier (\textit{5-level grayscale}), as well as including or not the Bayesian Localization step, with and without our proposed Branching-point Detector. Table~\ref{tab:results_ours} summarizes this experiment and demonstrates the contribution of all the components of our pipeline to the best final performance. 

On \textbf{phantom} data, our proposed pipeline achieves comparable results than using the baseline frame location classifier, with a slight improvement in sequence 0. 
More importantly, note the significant improvements in performance with \textbf{real data}. Our pipeline achieves the best top-1 and top-3 accuracy in the \textbf{real} sequences, with an average \textbf{22.5\%} and \textbf{32\%} improvement in top-1 and top-3 accuracy respectively, compared to using the original frame-based classifier and the basic Bayesian localization. Notably, the top-3 accuracy in both real sequences using our pipeline reaches similar results as with the phantom sequences.
More specifically, our proposed \textbf{frame-based classifier} improves the top-1 and top-3 accuracy on the real data by on average \textbf{8\%} and \textbf{6\%} compared to the original classifier, going from 31\% to 49\% top-1 accuracy on sequence Real 0 for example. On the other hand, the addition of the \textbf{branching-point detector} to the Bayesian localization step, when using our frame-based classifier, respectively leads to a further average \textbf{9.5\%} and \textbf{12.5\%} improvement of the top-1 and top-3 accuracy of the pipeline on real data. For example, in sequence Real 1, the top-1 accuracy is 37\% without the detector, while it reaches 49\% when the detector is added.

\begin{table}[!tb]
\caption{\textbf{Online Topological Localization} results, top-1 and top-3 accuracy (\textit{Acc.}), for variations of our proposed pipeline, including the existing baseline~\cite{keuth2024airway} for the classifier step. Results using phantom (\textit{Ph.}) and real data. }
\label{tab:results_ours}
\begin{tabular}{c c c c cc }
\toprule
 & \textbf{Frame} &  & \textbf{Branching} & \textbf{Top-1} & \textbf{Top-3}\\ 
\textbf{Seq.} & \textbf{Classifier} & \textbf{Bayesian} & \textbf{Detector} & \textbf{Acc.} & \textbf{Acc.} \\ 
\midrule
\multirow{6}{*}{Real}& \cite{keuth2024airway} & & & 0.31 & 0.53 \\
 & 5-level grayscale & & & 0.49 & 0.56 \\
 & \cite{keuth2024airway} & \checkmark & & 0.31 & 0.58 \\
 & 5-level grayscale & \checkmark & & 0.52 & 0.70 \\
0 & \cite{keuth2024airway} & \checkmark & \checkmark & 0.31 & 0.76 \\
 & 5-level grayscale & \checkmark & \checkmark & \textbf{0.59} & \textbf{0.91} \\
 \midrule
 
 \multirow{6}{*}{Real}& \cite{keuth2024airway} & & & 0.30 & 0.62 \\
 & 5-level grayscale & & & 0.28 & 0.71 \\
 & \cite{keuth2024airway} & \checkmark & & 0.32 & 0.59 \\
 & 5-level grayscale & \checkmark & & 0.37 & 0.86 \\
1 & \cite{keuth2024airway} & \checkmark & \checkmark & 0.39 & 0.62 \\
 & 5-level grayscale & \checkmark & \checkmark & \textbf{0.49} & \textbf{0.90} \\
 \midrule
 
 \multirow{6}{*}{Ph.0}& \cite{keuth2024airway} & & & 0.78 & 0.93 \\
 & 5-level grayscale & & & 0.82 & 0.95 \\
 & \cite{keuth2024airway} & \checkmark & & 0.78 & 0.91 \\
 & 5-level grayscale & \checkmark & & \textbf{0.83} & \textbf{0.96} \\
 & \cite{keuth2024airway} & \checkmark & \checkmark & 0.78 & 0.91 \\
 & 5-level grayscale & \checkmark & \checkmark & \textbf{0.83} & \textbf{0.96} \\
 \midrule
 
 \multirow{6}{*}{Ph.7}& \cite{keuth2024airway} & & & \textbf{0.63} & \textbf{0.84} \\
 & 5-level grayscale & & & \textbf{0.63} & 0.82 \\
 & \cite{keuth2024airway} & \checkmark & & 0.62 & 0.75\\
 & 5-level grayscale & \checkmark & & \textbf{0.63} & 0.80 \\
 & \cite{keuth2024airway} & \checkmark & \checkmark & 0.62 & 0.75 \\
 & 5-level grayscale & \checkmark & \checkmark & \textbf{0.63} & 0.80 \\
\bottomrule
\end{tabular}
\end{table}

Analyzing the limitations of our approach, the weakest component is the frame-based classifier. As it could be expected, it suffers from a lot of confusions due to the repetitive appearance of the captured scenes (see Fig.~\ref{fig:intro}), which results in relatively low accuracy and errors in the likelihood fed to the Bayesian filter. Although this is mitigated with the branching-point detector, the classifier may be improved with a more balanced and complete phantom dataset for training.

\section{Conclusions.}
 
This work presents the first online image based pipeline for topological localization during bronchoscopy procedures without the need for pre-operative CT scans, demonstrated in real data. 
Our pipeline consists of two steps, each capturing different aspects of the bronchoscopy images. The frame-based location classification model uses the structure and pixel intensities of each image to provide a prior of its localization, independently of the rest of the sequence. The Bayesian Localization model takes into account the temporal aspect of the bronchoscopy video feed and the tree structure of the airway to improve the individual frame classification. Our branching-point detector further improves the results by determining critical points in the localization, at branching-points of the tree when the scope is about to pass from one node to the next. 
We train our pipeline on phantom data only, and show promising results on real data labeled by experts. This avoids the long and costly process of collecting extensive real data from patients and having it manually labeled by experts. 
The presented pipeline could be adapted for other procedures but is particularly relevant in bronchoscopy due to the complex tree of the organ. 
Our results facilitate further efforts to build a complete navigational guidance system without relying on CT scans and registration techniques, and contribute towards the development of potential autonomous bronchoscopy systems in the future.

\section{Acknowledgments.}
 
This project has received funding from the European Union’s Horizon 2020 research and innovation programme under grant agreement No 863146 EndoMapper project, DGA project T45\_23R, and grants PID2021-125514NB-I00/MCIN/AEI/10.13039/501100011033/FEDER-UE and PID2022-139615OB-I00/MCIN/AEI/10.13039/501100011033/FEDER-UE.

\bibliographystyle{IEEEtran}
\bibliography{bibliography}

\begin{thebibliography}{10}
\providecommand{\url}[1]{#1}
\csname url@samestyle\endcsname
\providecommand{\newblock}{\relax}
\providecommand{\bibinfo}[2]{#2}
\providecommand{\BIBentrySTDinterwordspacing}{\spaceskip=0pt\relax}
\providecommand{\BIBentryALTinterwordstretchfactor}{4}
\providecommand{\BIBentryALTinterwordspacing}{\spaceskip=\fontdimen2\font plus
\BIBentryALTinterwordstretchfactor\fontdimen3\font minus \fontdimen4\font\relax}
\providecommand{\BIBforeignlanguage}[2]{{%
\expandafter\ifx\csname l@#1\endcsname\relax
\typeout{** WARNING: IEEEtran.bst: No hyphenation pattern has been}%
\typeout{** loaded for the language `#1'. Using the pattern for}%
\typeout{** the default language instead.}%
\else
\language=\csname l@#1\endcsname
\fi
#2}}
\providecommand{\BIBdecl}{\relax}
\BIBdecl

\bibitem{merritt2008image}
S.~A. Merritt, J.~D. Gibbs, K.-C. Yu, V.~Patel, L.~Rai, D.~C. Cornish, R.~Bascom, and W.~E. Higgins, ``Image-guided bronchoscopy for peripheral lung lesions: a phantom study,'' \emph{Chest}, vol. 134, no.~5, pp. 1017--1026, 2008.

\bibitem{memoli2012meta}
J.~S.~W. Memoli, P.~J. Nietert, and G.~A. Silvestri, ``Meta-analysis of guided bronchoscopy for the evaluation of the pulmonary nodule,'' \emph{Chest}, vol. 142, no.~2, pp. 385--393, 2012.

\bibitem{shinagawa2007virtual}
N.~Shinagawa, K.~Yamazaki, Y.~Onodera, F.~Asano, T.~Ishida, H.~Moriya, and M.~Nishimura, ``Virtual bronchoscopic navigation system shortens the examination time—feasibility study of virtual bronchoscopic navigation system,'' \emph{Lung Cancer}, vol.~56, no.~2, pp. 201--206, 2007.

\bibitem{reynisson2014navigated}
P.~J. Reynisson, H.~O. Leira, T.~N. Hernes, E.~F. Hofstad, M.~Scali, H.~Sorger, T.~Amundsen, F.~Lindseth, and T.~Lang{\o}, ``Navigated bronchoscopy: a technical review,'' \emph{Journal of bronchology \& interventional pulmonology}, vol.~21, no.~3, pp. 242--264, 2014.

\bibitem{ferguson2005virtual}
J.~S. Ferguson and G.~McLennan, ``Virtual bronchoscopy,'' \emph{Proceedings of the American Thoracic Society}, vol.~2, no.~6, pp. 488--491, 2005.

\bibitem{franz2014electromagnetic}
A.~M. Franz, T.~Haidegger, W.~Birkfellner, K.~Cleary, T.~M. Peters, and L.~Maier-Hein, ``Electromagnetic tracking in medicine—a review of technology, validation, and applications,'' \emph{IEEE Transactions on Medical Imaging}, vol.~33, no.~8, pp. 1702--1725, 2014.

\bibitem{mori2004new}
K.~Mori, T.~Enjoji, D.~Deguchi, T.~Kitasaka, Y.~Suenaga, J.~Toriwaki, H.~Takabatake, and H.~Natori, ``New image similarity measures for bronchoscope tracking based on image registration between virtual and real bronchoscopic images,'' in \emph{Medical Imaging 2004: Physiology, Function, and Structure from Medical Images}, vol. 5369.\hskip 1em plus 0.5em minus 0.4em\relax SPIE, 2004, pp. 165--176.

\bibitem{rai2008combined}
L.~Rai, J.~P. Helferty, and W.~E. Higgins, ``Combined video tracking and image-video registration for continuous bronchoscopic guidance,'' \emph{International Journal of Computer Assisted Radiology and Surgery}, vol.~3, pp. 315--329, 2008.

\bibitem{wang2020visual}
C.~Wang, M.~Oda, Y.~Hayashi, B.~Villard, T.~Kitasaka, H.~Takabatake, M.~Mori, H.~Honma, H.~Natori, and K.~Mori, ``A visual slam-based bronchoscope tracking scheme for bronchoscopic navigation,'' \emph{International Journal of Computer Assisted Radiology and Surgery}, vol.~15, pp. 1619--1630, 2020.

\bibitem{visentini2017deep}
M.~Visentini-Scarzanella, T.~Sugiura, T.~Kaneko, and S.~Koto, ``Deep monocular 3d reconstruction for assisted navigation in bronchoscopy,'' \emph{International journal of computer assisted radiology and surgery}, vol.~12, pp. 1089--1099, 2017.

\bibitem{shen2019context}
M.~Shen, Y.~Gu, N.~Liu, and G.-Z. Yang, ``Context-aware depth and pose estimation for bronchoscopic navigation,'' \emph{IEEE Robotics and Automation Letters}, vol.~4, no.~2, pp. 732--739, 2019.

\bibitem{banach2021visually}
A.~Banach, F.~King, F.~Masaki, H.~Tsukada, and N.~Hata, ``Visually navigated bronchoscopy using three cycle-consistent generative adversarial network for depth estimation,'' \emph{Medical image analysis}, vol.~73, p. 102164, 2021.

\bibitem{gil2020intraoperative}
D.~Gil, A.~Esteban-Lansaque, A.~Borras, E.~Ramirez, and C.~S. Ramos, ``Intraoperative extraction of airways anatomy in videobronchoscopy,'' \emph{IEEE access}, vol.~8, pp. 159\,696--159\,704, 2020.

\bibitem{esteban2016stable}
A.~Esteban-Lansaque, C.~S{\'a}nchez, A.~Borr{\`a}s, M.~Diez-Ferrer, A.~Rosell, and D.~Gil, ``Stable anatomical structure tracking for video-bronchoscopy navigation,'' in \emph{CLIP Workshop, MICCAI}.\hskip 1em plus 0.5em minus 0.4em\relax Springer, 2016, pp. 18--26.

\bibitem{sanchez2016navigation}
C.~S{\'a}nchez, M.~Diez-Ferrer, J.~Bernal, F.~J. S{\'a}nchez, A.~Rosell, and D.~Gil, ``Navigation path retrieval from videobronchoscopy using bronchial branches,'' in \emph{CLIP Workshop, MICCAI}.\hskip 1em plus 0.5em minus 0.4em\relax Springer, 2016, pp. 62--70.

\bibitem{yoo2021deep}
J.~Y. Yoo, S.~Y. Kang, J.~S. Park, Y.-J. Cho, S.~Y. Park, H.~I. Yoon, S.~J. Park, H.-G. Jeong, and T.~Kim, ``Deep learning for anatomical interpretation of video bronchoscopy images,'' \emph{Scientific Reports}, vol.~11, no.~1, p. 23765, 2021.

\bibitem{he2016identity}
K.~He, X.~Zhang, S.~Ren, and J.~Sun, ``Identity mappings in deep residual networks,'' in \emph{ECCV}.\hskip 1em plus 0.5em minus 0.4em\relax Springer, 2016, pp. 630--645.

\bibitem{huang2017densely}
G.~Huang, Z.~Liu, L.~Van Der~Maaten, and K.~Q. Weinberger, ``Densely connected convolutional networks,'' in \emph{CVPR}, 2017, pp. 4700--4708.

\bibitem{keuth2024airway}
R.~Keuth, M.~Heinrich, M.~Eichenlaub, and M.~Himstedt, ``Airway label prediction in video bronchoscopy: capturing temporal dependencies utilizing anatomical knowledge,'' \emph{International Journal of Computer Assisted Radiology and Surgery}, vol.~19, no.~4, pp. 713--721, 2024.

\bibitem{smith2018human}
B.~M. Smith, H.~Traboulsi, J.~H. Austin, A.~Manichaikul, E.~A. Hoffman, E.~R. Bleecker, W.~V. Cardoso, C.~Cooper, D.~J. Couper, S.~M. Dashnaw \emph{et~al.}, ``Human airway branch variation and chronic obstructive pulmonary disease,'' \emph{Proceedings of the National Academy of Sciences}, vol. 115, no.~5, pp. E974--E981, 2018.

\bibitem{keuth2023weakly}
R.~Keuth, M.~Heinrich, M.~Eichenlaub, and M.~Himstedt, ``Weakly supervised airway orifice segmentation in video bronchoscopy,'' in \emph{Medical Imaging}, vol. 12464.\hskip 1em plus 0.5em minus 0.4em\relax SPIE, 2023, pp. 66--73.

\bibitem{morlana2024colonmapper}
J.~Morlana, J.~D. Tard{\'o}s, and J.~Montiel, ``Colonmapper: topological mapping and localization for colonoscopy,'' in \emph{ICRA}.\hskip 1em plus 0.5em minus 0.4em\relax IEEE, 2024, pp. 6329--6336.

\bibitem{thrun2005probabilistic}
S.~Thrun, W.~Burgard, and D.~Fox, \emph{Probabilistic Robotics}.\hskip 1em plus 0.5em minus 0.4em\relax MIT Press, 2005.

\bibitem{cubuk2019autoaugment}
E.~D. Cubuk, B.~Zoph, D.~Mane, V.~Vasudevan, and Q.~V. Le, ``Autoaugment: Learning augmentation strategies from data,'' in \emph{CVPR}, 2019, pp. 113--123.

\end{thebibliography}

\end{document}